\documentclass{article}

\usepackage{arxiv}

\usepackage[utf8]{inputenc} 
\usepackage[T1]{fontenc}    
\usepackage{hyperref}       
\usepackage{url}            
\usepackage{booktabs}       
\usepackage{amsfonts}       
\usepackage{nicefrac}       
\usepackage{microtype}      
\usepackage{lipsum}         
\usepackage{graphicx}
\usepackage{doi}
\usepackage{mathtools}
\usepackage[table,xcdraw]{xcolor}
\usepackage{multirow}
\usepackage{amsmath}
\usepackage{cleveref}       
\usepackage{bbm}
\usepackage{algorithm}
\usepackage{algorithmic}

\title{Multi-View Contrastive Learning from Demonstrations}

\date{}

\author{ \hspace{1mm}André Correia \\
	NOVA LINCS\\
	Universidade da Beira Interior\\
	Covilhã, Portugal \\
	\texttt{andre.correia@ubi.pt} \\
	\And
	\hspace{1mm}Luís A. Alexandre \\
	NOVA LINCS\\
	Universidade da Beira Interior\\
	Covilhã, Portugal \\
	\texttt{luis.alexandre@ubi.pt} \\
}

\renewcommand{\shorttitle}{Multi-View Contrastive Learning from Demonstrations}

\hypersetup{
pdftitle={Contrastive Learning from Demonstrations},
pdfsubject={q-bio.NC, q-bio.QM},
pdfauthor={André Correia, Luís A. Alexandre},
pdfkeywords={Contrastive, Learning, from, Demonstrations},
}

\begin{document}
\maketitle

\begin{abstract}
This paper presents a framework for learning visual representations from unlabeled video demonstrations captured from multiple viewpoints. We show that these representations are applicable for imitating robotic tasks. We use contrastive learning to enhance task-relevant information while suppressing irrelevant information in the feature embeddings.
We validate the proposed method on the publicly available Multi-View Pouring and a custom Pick and Place data sets and compare it with the TCN and CMC baselines. We evaluate the learned representations using three metrics: viewpoint alignment, stage classification and reinforcement learning. In all cases, the results improve when compared to state-of-the-art approaches.
\end{abstract}


\begin{figure}[!tb]
\centering
  \begin{minipage}{\linewidth}
    \makebox[\linewidth]{\includegraphics[width=0.5\linewidth]{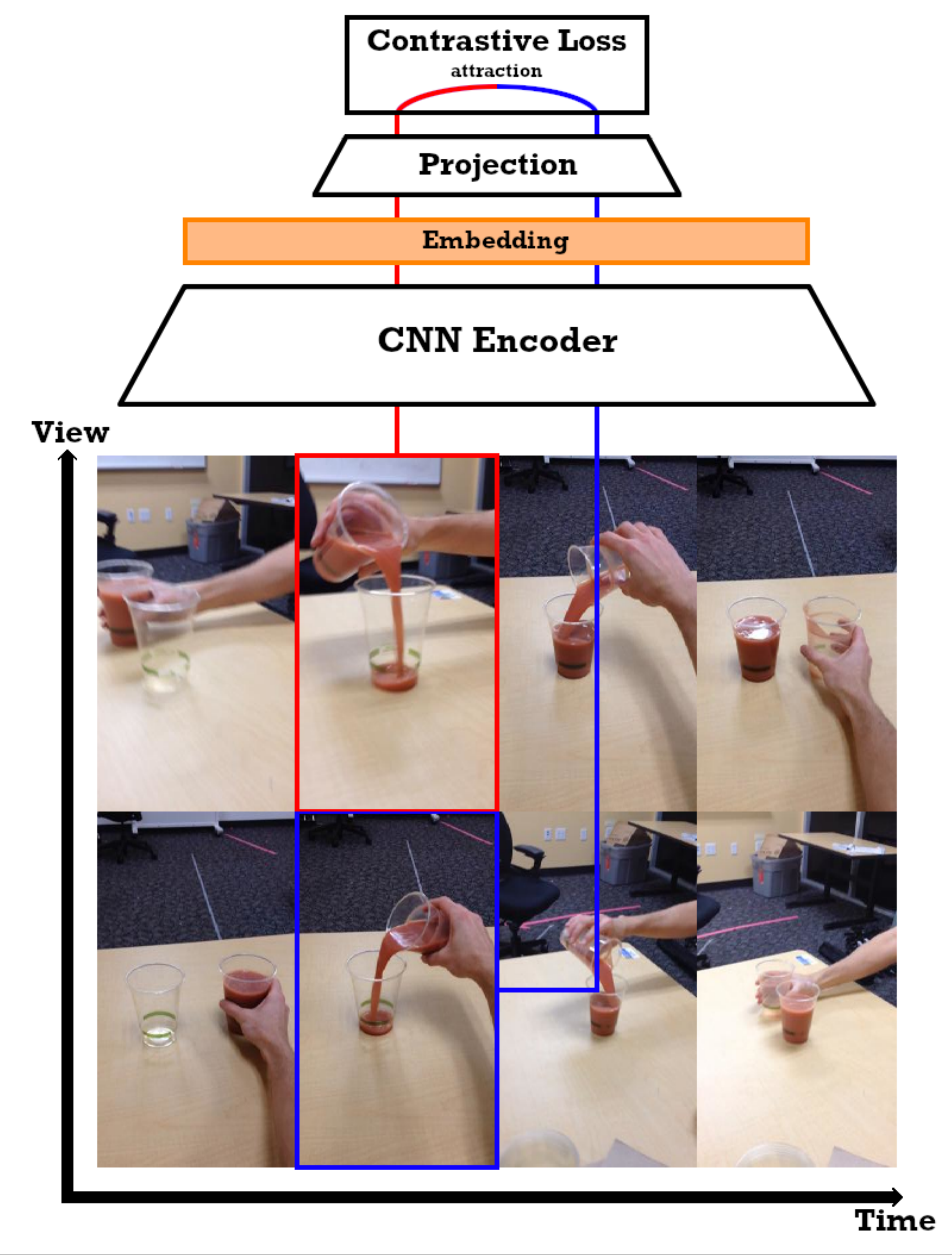}}%
  \end{minipage}
  \caption{CLfD framework: The data set of synchronized multi-view video pairs generates a set of anchor-positive image pairs. The CNN encoder generates the respective feature embedding pairs. A projection network maps the embeddings to the learning space where the loss is applied. The networks' parameters are updated to maximize the agreement between the pair of features using the contrastive loss. The embeddings can replace the images for viewpoint invariant representations in various tasks.}\label{Fig:MV}
\end{figure}

\section{Introduction}
Reinforcement learning (RL) and demonstration learning (DL) algorithms have enabled human-robot collaboration. 
Although these methods reach a working policy and are computationally efficient, the set of visited environment states present in the demonstrations is small, and they do not generalize to scenarios unseen during the demonstration phase.
Because of this, DL approaches are frequently coupled with RL to extend the agent's knowledge of the environment and consequently increase its generalization capability. 

However, such policies are trained under a static domain and will likely struggle to perform in other domains. Minor changes to the positions of the cameras, illumination, and objects in the scene change the content of the images representing the state of the task. Ideally, robots would extract task-relevant attributes from the images while ignoring the irrelevant noise information also present in the images.
Existing solutions include supervised learning for tasks where humans can easily specify labels, such as object classification. Examples of possible attributes could be background, containers, and viewpoint. However, most robotic tasks are too complex, and it becomes impractical to label every attribute for every state in the demonstration. 
Other solutions make use of metric learning, which is computationally complex and makes the learning and inference processes slow \cite{facenet}.
We alleviate these issues through the means of self-supervised and Contrastive Learning (CL) \cite{contrastive1}. We train an encoder on multi-viewpoint video demonstrations of the robotic task, where time is the supervision signal. CL encodes the state images into viewpoint-invariant representations. 

This paper presents a demonstration learning framework for extracting image embeddings that enhance task-specific information while suppressing task-irrelevant noise. We show that the representations can be applied to learn robotic tasks through standard RL algorithms using a single demonstration to provide the reward signal. The proposed framework does not require any prior knowledge about the task beforehand. We validate our method on the publicly available Multi-View Pouring data set and a custom Pick and Place simulation environment and data sets.
The proposed model reaches equal performance while requiring less training time when compared with other baselines. Additionally, ablation studies were performed on different components of the method to identify the best configuration and the contributions of the individual components.

The main contributions made by this paper are the multi-view Contrastive Learning from Demonstrations (CLfD) framework, which contrastively learns viewpoint invariant representations from demonstrations, the Pick and Place task environment in CoppeliaSim and the respective multi-view demonstration data set for bench-marking RL and DL methods.
The rest of this paper is organized as follows. Section \ref{sec:relatedwork} provides a brief overview of the state-of-the-art. The proposed approach is explained in section \ref{sec:proposed}. Experiments and corresponding results are presented and discussed in section \ref{sec:experiments}. Lastly, in section \ref{sec:conclusions} the conclusions are drawn.

\section{Related Work}
\label{sec:relatedwork}
Demonstration learning is a machine learning paradigm that allows robots to learn tasks from demonstrations performed by human experts and has been widely studied in the field of robotics \cite{senhoras,SchaalBillard,recentadv}.
The two predominant areas of demonstration learning are behavioural cloning (BC) and inverse reinforcement learning (IRL).

BC treats demonstration learning as a supervised learning problem, where the problem maps observations into actions, and the training signal is given by how similar the actions are to the demonstrator's \cite{mario}. BC struggles to generalize to unseen observations due to the difficulty of collecting actions in demonstrations.
On the other hand, IRL treats DL as a RL problem, alleviating the previous problem by collecting interaction data. It uses DL to define the reward function used to estimate the policy \cite{shaping,shaping2,brys}.
Both types of demonstration learning typically require the expert demonstrator to be in the same context as the learner, limiting their scalability to real-world applications. Moreover, applications that train the learner in the same context as the demonstrator rely on kinesthetic \cite{kinesthetic,jointspace} or teleoperation \cite{NASA,helicopter,soccer} demonstration techniques. However, these require expert skills to perform and still limit the observation setting to the one in the demonstration.

Several works have tackled the problem of learning from demonstrations captured from multiple contexts \cite{multipleviews1}. The context changes can vary from viewpoints, backgrounds, illumination, or agents.
In \cite{gupta,crosscontext}, a translation of the recorded observations from the demonstrator's context to the context of the learner is used. However, their approach only works for simple tasks which can be represented solely by the first and the final observations.
Recently \cite{squirl} encoded a demonstration and provided the embedding as an extra signal for the policy to perform in a novel context. However, this requires a demonstration for the given context to be available and user interaction to define which demonstration to encode.

Unsupervised learning has been used to learn policies from unlabelled visual demonstrations. Previous methods have made use of multiple modalities to obtain rich embeddings.
In \cite{tcc} an encoder is trained to estimate similar features for concurrent frames of multi-view synchronized videos. Here the criterion is cycle-consistency, where for two views, a data point is cycle-consistent if the nearest neighbour of its nearest neighbour is the point itself. 

Triplet learning has been used to estimate an embedding space where similar observations are closer in space than time-distant ones.
This method was applied to demonstration learning using Time Contrastive Networks (TCN) \cite{tcn}, where synchronized multi-view demonstration videos are used to estimate an embedding space where observations from different viewpoints but identical timestamps generate identical features.
Later, in \cite{atnet} the encoder was enhanced with multiple levels of attention to further focus on viewpoint invariant features.
In \cite{disentagled}, they explicitly disentangle state and viewpoint features from the feature vector using an encoder-decoder architecture combined with a permutation loss. 
The main drawback of the previous methods is how computationally intensive they are to train because they rely on organizing the data set into triplets (sets of anchor, positive and negative images). 
Sampling a negative image for an anchor-positive pair is not trivial. On the one hand, sampling every possible combination leads to an extensive data set to process and will include weak negatives (the anchor-negative distance is easily greater than the anchor-positive distance).
On the other hand, sampling hard negatives (anchor-negative distance is close to the anchor-positive distance) is a complex task that comes with downsides such as the frequent evaluation and re-sampling of negatives \cite{hardsoft}.

CL compares different images sharing a common signal to learn representations in a self-supervised fashion.
It has been applied to multiple machine learning fields, most notably image classification \cite{simclr}, where the embedding space is robustly obtained in a self-supervised manner by bringing images from the same class closer in the space. 
Siamese networks \cite{towardssiamese} have been paired with CL where each network outputs the features of a different view. In \cite{towardssiamese}, the different views are obtained through data augmentation, and applied to motion simulation tasks.

Other ways to obtain different views from a single image are by using different channels. In Contrastive Multiview Coding (CMC) \cite{contrastivecoding}, the images are converted into the Lab colour space, where the L and ab components are treated as two views of the image. Then contrastive loss is applied to learn an embedding space. Additionally, they show that increasing the number of views leads to better features.


\section{Proposed Approach}
\label{sec:proposed}

Our method applies CL \cite{contrastive1} to multi-view demonstration learning and is represented in Fig. \ref{Fig:MV}.
CLfD learns viewpoint-independent state representations by maximizing the agreement between synchronized frames from different viewpoints in the embedding space using the contrastive loss. We assume to have access to a demonstration data set where each demonstration is captured from varying viewpoints and the frames of the different views are synchronized. Frames from the same demonstration and timestamp but different viewpoints represent the same world state, we call these anchor-positive pairs. We want to obtain representations that capture task-relevant features and ignore task-irrelevant ones. We do this by encouraging the features from anchor-positive pairs to be close in the embedding space through CL. Our framework is described in Algorithm \ref{alg:clfd} and is composed of three modules.

A deep convolutional neural network encoder extracts the frames' embedding vectors. The framework is flexible and allows for changes to the network without requiring changes to the remaining modules, as long as it ends with a fully connected layer outputting $1 \times N$ feature vectors, where $N$ is the number of features each embedding vector should contain. We use vectors with 32 features. We tested several network architectures to act as the encoder. We present the results in section \ref{sec:experiments}.
The authors \cite{simclr} found that using an additional small nonlinear projection neural network was more beneficial than applying the contrastive loss directly to the features.
Therefore, we also used an MLP with a single hidden layer, with $N$ neurons, as the projection network to project the features obtained from the encoder to the space where the contrastive loss is applied. 
We project the features to 64-dimensional vectors. 

Lastly, we apply the contrastive loss introduced in \cite{simclr} to the output of the projection head. Originally the authors applied a set of transformations to generate a positive pair from the anchor image. The multi-view data set directly provides the pairs. We use this loss because it applies to contrastive prediction tasks without a prior sampling of negative examples. Given a batch of anchor-positive pairs of size $N$, $\{x_{ki}, x_{kj}\}$ where $0 \leq k < N$, the contrastive loss aims to pair each anchor element $x_i$ with its respective positive pair $x_j$ in the full set of elements $\{x_{1...2N}\}$. Instead of explicitly sampling negative examples such as in triplet learning, for each anchor image $x_i$, the remaining $2(N-1)$ images in the batch of size $N$ pairs act as negative examples relative to the anchor.
Not having to explicitly sample negative pairs corresponds to a significant complexity and performance advantage over methods that require it.

\begin{algorithm}[tb]
   \caption{CLfD algorithm}
   \label{alg:clfd}
\begin{algorithmic}
    \STATE {\bfseries Input:} Contrastive data set $D = \{x_a, x_p\}$, batch size $N$, training iterations $I$, constant $\tau$, encoder $f$, projector $g$.
    \FOR{$i=1$ {\bfseries to} $I$}
        \FOR{$batch$ $\{x_{a_{t:N}}, x_{p_{t:N}}\}$ {\bfseries in} $D$}
            \STATE $h_{a_{t:N}} = f(x_{a_{t:N}})$ 
            \STATE $h_{p_{t:N}} = f(x_{p_{t:N}})$ 
            \STATE $z_{a_{t:N}} = g(h_{a_{t:N}})$ 
            \STATE $z_{p_{t:N}} = g(h_{p_{t:N}})$ 
            \FORALL{$i \in \{1, ..., 2N\}$ and $j \in \{1, ..., 2N\}$}
                \STATE $s_{i,j} = z_i^Tz_j/(\|z_i\| \|z_j\|)$
            \ENDFOR
            \STATE $\mathcal{L} = \frac{1}{2N}\sum\limits_{k=1}^{N} [\ell(2k-1, 2k) + \ell(2k, 2k-1)]$
            \STATE Update $f$ and $g$ to minimize $\mathcal{L}$ using Adam optimizer.
        \ENDFOR
    \ENDFOR
    \STATE {\bfseries return} $f$
\end{algorithmic}
\end{algorithm}

The loss function for a pair of anchor-positive examples $(x_a, x_p)$ is \cite{simclr}:

\begin{equation}
    \ell(x_a, x_p) = -log \frac{exp(sim(z_a, z_p)/ \tau)}{\sum\limits_{k=1}^{2N} \mathbbm{1}_{[k \neq a]} exp(sim(z_a, z_k)/\tau)}
\end{equation}
where $\mathbbm{1}_{[k \neq a]} \in \{0,1\}$ is 1 if $k \neq a$ otherwise is zero, $\tau$ is a temperature hyper-parameter which we set to 0.5, and $sim(a,b)$ is the cosine similarity between two vectors. Lastly, the final loss $\mathcal{L}$ used to update the parameters of the networks is the normalized temperature-scaled cross-entropy loss (\textit{NT-Xent}) \cite{simclr} computed using the pair losses across the batch.

\begin{figure}[!tb]
\centering
  \begin{minipage}{\linewidth}
    \makebox[\linewidth]{\includegraphics[width=0.5\linewidth]{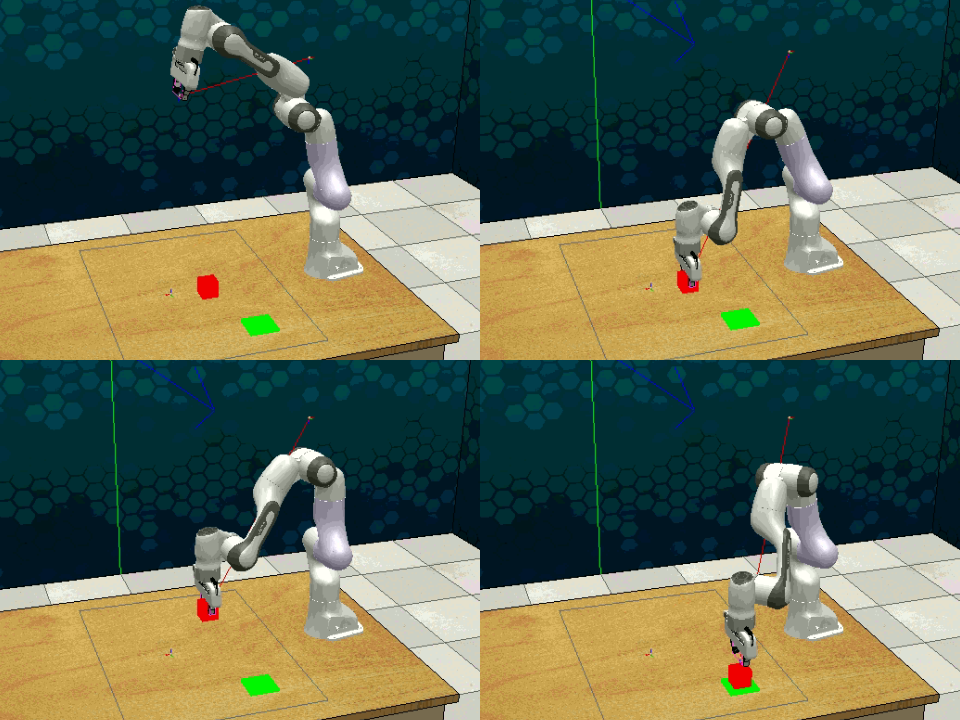}}%
  \end{minipage}
  \caption{Screenshot of the pick and place task environment in CoppeliaSim \cite{coppelia}. The 7-DOF simulated Panda arm must first place its end-effector near the red box and close the gripper to pick it up. Then the robot must move its end-effector above the goal position marked by a green plane and open the gripper, causing the box to fall on top of the plane. Images are simultaneously captured from 5 viewpoints: first-person (camera attached to the gripper), front, top, overhead, and right-side viewpoints. We use the environment to capture the demonstration data set and to train the DDPG \cite{ddpg} agent. In each new demonstration, the locations of the box and the green plane are randomly changed.}\label{figenvironment}
\end{figure}

\subsection{Pick and Place Multi-View Data Set}

\begin{figure}[!tb]
\centering
  \begin{minipage}{\linewidth}
    \makebox[\linewidth]{\includegraphics[width=0.5\linewidth]{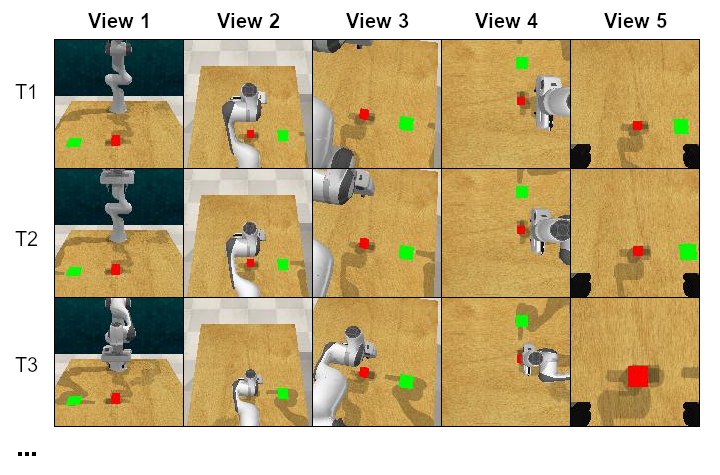}}%
  \end{minipage}
  \caption{Examples of stored views in the custom demonstration data set for the Pick and Place task simulated in CoppeliaSim \cite{coppelia}, captured from 3 timestamps. Each demonstration is captured from 5 different fixed viewpoints.}\label{figdataset}
\end{figure}

We created a custom Pick and Place task environment in CoppeliaSim \cite{coppelia}.
The environment is composed of a 7-DOF Panda arm on top of a table. The arm must pick up the red box and place it on the goal location marked by the green plane. Additionally, there are five fixed cameras capturing observations from different viewpoints: one attached to the robot, one in front, another above, one behind,  and one on the right side of the robot. The environment is represented in Fig.~\ref{figdataset}.

Using this environment, we then created a custom multi-view demonstration data set for the pick-and-place task. 
The demonstrations are performed automatically by having access to the position of the objects inside the simulation. With them, we obtain the joint paths to pick and to place the box, using inverse-kinematics. We perform the path, and after each transition, we obtain the images from each camera, as well as the robot's current joint angles and velocities and store them in the data set.
The initial positions of the box and stack are randomized in each demonstration. 

The generated data set is composed of 150 demonstrations, a realistic size, smaller than what is used in the literature \cite{tcn,squirl,atnet}, due to the difficulty of obtaining real demonstrations.
We select 100 demonstrations for training, 25 for validation, and the remaining 25 for testing. Unlike \cite{atnet}, we do not include failed demonstrations in our data set. Some examples of views stored in the data set are presented in Fig.~\ref{figdataset}.
This data set and our code are available at https://tinyurl.com/mvclfdemos.

\subsection{Intuition towards using CL for Demonstration Learning}
Triplet learning has been successfully combined with demonstration learning to estimate a viewpoint invariant space by pulling anchor-positive images closer while repelling anchor-negative images in the space. The embeddings can then be applied to pose demonstration and reinforcement learning tasks \cite{tcn}. However, it is a slow and complex method. It requires three images and the sampling and frequent re-sampling of negative examples for each anchor-positive pair.

CL has been applied to more traditional machine learning approaches, such as object classification \cite{simclr}. Typically the positive image is obtained from the anchor image through transformations. Alternatively, in \cite{contrastivecoding} the images are converted into Lab colour space and the anchor-positive images correspond to each channel. Then, the embedding space is estimated by pulling the features from the generated images closer to the original image or another image generated from a different set of transformations.

We avoid the downsides of triplet learning  by replacing it with CL. Using synchronized video pairs, we avoid the need to generate fake images and since both images are authentic, we ensure they are contrasting in the viewpoint, causing the model to converge much faster and the embedding space to be viewpoint invariant. Additionally, by using CL, we reduce the number of images from three to two compared to triplet learning. Consequently, removing the need to sample negative examples reduces complexity and fastens the learning process. We use the (\textit{NT-Xent}) contrastive loss to repel the features from images outside the anchor-positive pair in the embedding space, similarly to triplet learning.

\subsection{Using the features for RL}
\label{sec:rl}
We show that the features obtained from our model can be used to learn a policy through RL. The agent learns to pick up a box and place it on top of a plane using the embedding obtained from the encoder as the state representations. The agent learns the task using a single demonstration for guidance. The demonstration used for RL was not used during training.

The RL problem is a standard Markov Decision Process (MDP), defined by the tuple $<S, A, R, T, \gamma>$ \cite{barto}, where $S$ is the set of states, $A$ is the set of actions, $R$ is the reward function, $T$ is the state transition function, and $\gamma$ is the discount factor.
Each state is composed of the embedding vector representation of the state image obtained from the encoder along with the robot's joint angles and velocities and binary gripper state. This ensures representation of both the environment and robot states. The actions are the joint velocities and gripper state. 

For multi-stage tasks, such as the Pick and Place task, we treat each stage as an individual task and train a policy for each stage to speed up convergence. The reward at timestep $t$ is defined by how close the environment's state embeddings are to the last demonstration state embeddings for the respective stage: 
\begin{equation}
    R(t) = - \|f({s_e}_t) - f({s_d}_g)\|
\end{equation}
where $f({s_e}_t)$ is the embedding of environment's current state, while $f({s_d}_g)$ is the last state embedding for the stage $g$ in the demonstration. We set the discount factor $\gamma$ to 0.99.
We use an off-the-shelf RL algorithm, DDPG \cite{ddpg}, to estimate the policy because it is compatible with continuous action spaces and enhances it with HER \cite{her} to speed up convergence.

\section{Experiments}
\label{sec:experiments}

In this section, we define the methods used to evaluate the framework, which helps to understand the different design choices performed. We performed the experiments on a machine equipped with an Intel i7 980 with 3.33 GHz and 6 cores, 24 GB of RAM, and an Nvidia GTX 1080Ti GPU. We train the models using the Pytorch framework \cite{pytorch} and use its builtin Resnet18 implementation. Use CMC \cite{contrastivecoding} and TCN \cite{tcn} baseline implementations provided in their respective works.

\subsection{Alignment Error}
\begin{table}[tb]
\centering
\caption{Comparison between the CLfD method with the TCN \cite{tcn} and CMC \cite{contrastivecoding} baselines using the alignment error percentage metric, varying the encoding architecture and data set. Models were trained for 1000 epochs and using a batch size of 50. \label{tab:algerr}}
\begin{tabular}{c|c|ccc}
\multirow{2}{*}{Data Set}  & \multirow{2}{*}{Encoder} & \multicolumn{3}{c}{Alignment Error {[}\%{]}}                                        \\ \cline{3-5} 
                           &                          & \multicolumn{1}{c|}{TCN} & \multicolumn{1}{c|}{CMC}            & CLfD           \\ \hline
\multirow{2}{*}{PickPlace} & Resnet18                 & \multicolumn{1}{c|}{24.98}   & \multicolumn{1}{c|}{\textbf{12.79}} & 16.82          \\
                           & TCN                      & \multicolumn{1}{c|}{15.35}   & \multicolumn{1}{c|}{4.45}           & \textbf{1.90}  \\ \hline
\multirow{2}{*}{Pouring}   & Resnet18                 & \multicolumn{1}{c|}{27.20}   & \multicolumn{1}{c|}{27.80}          & \textbf{23.28} \\
                           & TCN                      & \multicolumn{1}{c|}{26.74}   & \multicolumn{1}{c|}{24.57}          & \textbf{19.84}
\end{tabular}
\end{table}

We compare the performance of our method with the TCN \cite{tcn} and CMC \cite{contrastivecoding} baselines using the alignment error metric, proposed in \cite{tcn}. For the TCN baseline, the negative pairs are randomly sampled. For the CMC baseline, we don't perform the conversion into the Lab colour space because we can directly obtain the anchor-positive pairs from the data set. The alignment error metric measures how well the model aligns the frames of synchronized video pairs based on their embeddings. Simultaneously, it evaluates how well the model approximates embeddings from concurrent frames and repels non-concurrent embeddings. We test two different encoders on the two data sets.
The alignment error between two synchronized videos is:
\begin{equation}
    AE(v_1, v_2) = \sum\limits_{i=1}^{N} \underset{j}{\mathrm{min}}(\|v_{1i} - v_{2j}\|) / N
\end{equation}
where $v_{ki}$ is the $i$-th frame of video $v_k$, and $N$ is the total number of frames in each video.
First, we evaluate the framework on the publicly available Multi-View Pouring data set \cite{tcn}, which contains 235 synchronized video pairs of a person demonstrating pouring liquids from one container to another. In each pair, one video is captured from a static viewpoint while the other is moved around the task, capturing different viewpoints.
Additionally, we used the custom demonstration data set for the pick-and-place task simulated in CoppeliaSim \cite{coppelia}.

The methods are trained on the data sets for 1000 epochs using batches of 32. All encoders are pre-trained on the ImageNet \cite{imagenet} data set. We calculated the alignment error of the model using the validation set every 25 epochs and present the lowest value obtained throughout training. We compare the performance of the Resnet18 encoder with the TCN encoder from the baseline \cite{tcn}. 
It is clear from Table \ref{tab:algerr} that CLfD outperforms the TCN baseline in all four setups.
With the custom Pick-and-Place data set, CLfD reduces the alignment error by over 8\% using Resnet18 encoder and 13.45\% using the TCN encoder compared with the TCN baseline. For the real-world multi-view pouring data set, CLfD reduces the alignment error by nearly 4\% using the Resnet18 encoder and by 6.9\% using the TCN encoder.

CLfD outperforms the CMC baseline in three out of four setups. CLfD shows an alignment error lower than CMC by 2.55\% when using TCN encoder, for the Pick and Place data set, and by 4.52\% and 4.73\%, when using Resnet18 and TCN encoders respectively, for the Pouring data set. CMC does outperform our method in the Pick and Place data set when using Resnet18 encoder. However, we obtain a smaller alignment error by using the TCN encoder.
In fact, the TCN encoder outperforms Resnet18 in all setups, proving that it can also be used in CL to encode viewpoint invariant features.

Lastly, we obtain an alignment error near the value of 18.8\% for the multi-view Pouring data set, which is the one presented by the baseline in \cite{tcn}. However, this TCN baseline encoder is trained on the multi-view pouring data set for 397 thousand iterations while sampling negative examples. Therefore, our model converges in far fewer epochs, with faster training and without a specialized algorithm for sampling the data.

\subsection{Stage Classification}

\begin{table}[tb]
\centering
\caption{Classification accuracy, for determining the stage of the demonstration, is obtained from evaluating an MLP with two fully connected layers on the Pick or Place classification data set's test set. The MLP is trained over the features obtained from a TCN encoder trained using the CLfD algorithm. The accuracy is calculated for camera angles seen and unseen camera viewpoints during the training of the TCN encoder. The MLP obtains near-perfect accuracy proving that the features are viewpoint independent. \label{tab:stage}}
\begin{tabular}{l|c}
\multicolumn{1}{c|}{ViewPoints} & Accuracy {[}\%{]} \\ \hline
Seen                             & 100               \\
Seen + Unseen                    & 99.41             \\
Unseen                           & 98.69            
\end{tabular}
\end{table}

We use a variant of the labelled classification error as seen in \cite{tcn,atnet}, which requires labelling different features for each frame. Instead, we classify the stage of the task from the embeddings. The stage classification metric measures how well the model can disentangle stage-related attributes from the features, proving that they contain valuable information to be used in a real robotic task. We reorganised the Pick-and-Place demonstration data set into a classification data set. The new data set is organized into two classes, pick and place. Each class contains the exact number of images, totalling 4000 examples. The images are evenly distributed amongst the five cameras. A TCN encoder is first trained on the multi-view Pick-and-Place data set using the CLfD algorithm. The encoder generates the features from the images of the classification data set. The features are forwarded to an MLP with two fully connected layers. The MLP is trained for 500 epochs. We evaluate the accuracy of the MLP for predicting the correct stage from the features using 200 unseen test images.

We also test whether the features are viewpoint independent by comparing the MLP's performance when trained on viewpoints not seen by the encoder during its training. The encoder was trained on images captured by three specific cameras. We refer to these as seen viewpoints. The images captured by the remaining two cameras are from unseen viewpoints. Table \ref{tab:stage} shows that the MLP obtains near-perfect results even for images captured from unseen viewpoints during the training of the encoder. These results show that the encoded features are viewpoint independent and therefore can be used in real robotic tasks to encode the state image into a feature vector regardless of the viewpoint from where the image was captured.

\subsection{Reinforcement Learning}

\begin{figure}[!tb]
\centering
  \begin{minipage}{\linewidth}
    \makebox[\linewidth]{\includegraphics[width=0.5\linewidth]{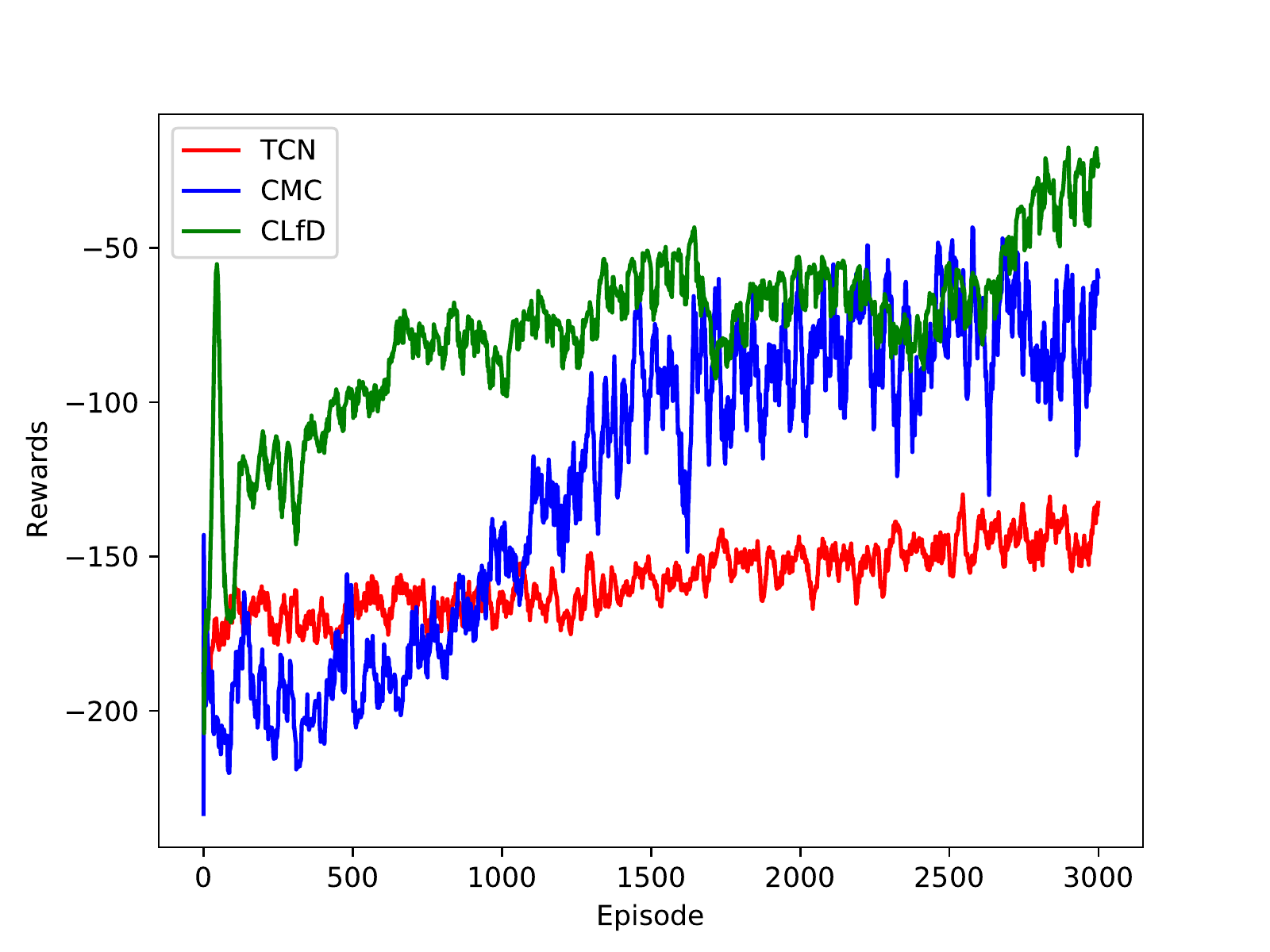}}%
  \end{minipage}
  \caption{Accumulated reward by the DDPG \cite{ddpg} agent over 3000 episodes when trained using
CLfD, CMC and TCN encoders with TCN network \cite{tcn} as the backbone for learning the Pick stage of the Pick and Place task in the simulated environment.}\label{fig:rl}
\end{figure}

Lastly, we determine whether the embedded features can replace the state as viewpoint invariant representations for reinforcement learning tasks as well as if they can be used to provide rewards, by evaluating the accumulated rewards over time. 
The RL agent described in section \ref{sec:rl} is trained using the CLfD encoder with the TCN architecture as the backbone, trained on the Pick-and-Place data set. The agent must solve the Pick and Place task. Because it is a two-stage task and we're using the DDPG algorithm with sparse rewards, we split each stage of the task into individual tasks for faster convergence. The agent must learn a policy for each stage. The agent is trained for 3000 episodes, after which the agent learns each stage. The accumulated reward after each episode, for the pick stage, is calculated and shown in Fig.~\ref{fig:rl}. An identical graph is obtained for the place stage. 

We compare the performance of the agent using the CLfD encoder's features with the performance obtained when using the features obtained from the encoder trained with the TCN and CMC baselines. The agent is not able to learn the task when using the features of the baselines. Particularly, the TCN agent still performs random actions after the 3000 episodes. CMC gets close to picking up the box and to placing the box but fails by a small misposition of the gripper. The results show that: the performance of TCN \cite{tcn} is dependent on the efficient sampling of negative pairs and that CLfD's features can be used to learn this type of task with a smaller computational effort.
This further cements that the proposed framework is able to learn an embedding space capable of representing the environment's state and that such representations can be used directly as a replacement of the state image for the agent's input as well as for guiding the agent through the learning process by providing the rewards.

\section{Conclusions}
\label{sec:conclusions}

This work presents a framework that uses contrastive learning to obtain viewpoint-invariant state representations from demonstrations.
The representations are obtained by contrasting semantically aligned frames from different viewpoints. The semantic alignment is ensured by synchronizing video demonstrations, as is done in \cite{tcn,atnet}.

We carefully study different backbones for the encoders and show the effects of different design choices. We show that our framework can correctly align frames between two viewpoints and with fewer training iterations than triplet learning while being lighter to compute and more straightforward to organize the data. We also created a demonstration learning data set that can be used to explore other approaches and compare them against our proposal.
Lastly, we showed that the representations contain information that can be used for classification purposes and provide a reward function within a reinforcement learning algorithm to learn a manipulation task.

\section*{Acknowledgements}
This work was supported by NOVA LINCS (UIDB/04516/2020) with the financial support of ‘FCT - Fundação para a Ciência e Tecnologia’ and also through the research grant ‘2022.14197.BD’.

\end{document}